\documentclass[runningheads]{llncs}
\usepackage{graphicx}

\usepackage{tikz}
\usepackage{comment}
\usepackage{amsmath,amssymb} 
\usepackage{color}
\usepackage{orcidlink}

\usepackage[accsupp]{axessibility}  

\usepackage{graphicx}
\usepackage{algorithm}
\usepackage{bm}
\usepackage{color}
\usepackage{xcolor}
\usepackage{multirow}
\usepackage{ifthen}
\usepackage{bbm}
\usepackage{pifont}

\graphicspath{{./figures/}}

\def\onedot{.}
 
\def\ie{\emph{i.e}\onedot} 
 
\def\etc{\emph{etc}\onedot} \def\vs{\emph{vs}\onedot}
\def\wrt{w.r.t\onedot} 
 
\def\etal{\emph{et al}\onedot}

\DeclareMathOperator{\agg}{\text{MaxPool}}
\DeclareMathOperator{\fc}{\text{FC}}
\DeclareMathOperator{\softmax}{\text{Softmax}}
\DeclareMathOperator{\bce}{\text{BCE}}
\newcommand{\keypoint}[1]{\noindent\textbf{#1}\quad}

\newcommand{\method}[1]{\ifthenelse{\equal{#1}{full}}{Elastic Moment Bounding}{EMB}}

\newcommand{\best}[1]{{{\color{red}#1}}}
\newcommand{\scnd}[1]{{{\color{blue}#1}}}

\newcommand{\jh}[1]{{\color{black}#1}}

\begin{document}
\pagestyle{headings}
\mainmatter
\def\ECCVSubNumber{2818}  

\title{Video Activity Localisation with \\
Uncertainties in Temporal Boundary} 

\titlerunning{Elastic Moment Bounding}
%
\author{Jiabo Huang\inst{1,4} \and
Hailin Jin\inst{2} \and
Shaogang Gong\inst{1} \and
Yang Liu\thanks{Corresponding author.}\inst{3,5}
}
\authorrunning{J. Huang et al.}
%
\institute{
Queen Mary University of London \\\email{\{jiabo.huang, s.gong\}@qmul.ac.uk} \\
\and Adobe Research \\\email{hljin@adobe.com} \\
\and Wangxuan Institute of Computer Technology, Peking University \\\email{yangliu@pku.edu.cn}
\and Vision Semantics Limited
\and Beijing Institute for General Artificial Intelligence
}

\maketitle

\begin{abstract}
Current methods for video activity localisation over time
assume implicitly that activity temporal boundaries labelled for model
training are determined and precise.  However, in unscripted natural videos, different activities mostly transit smoothly, so that it is intrinsically ambiguous to determine in labelling precisely when an activity starts and ends over time. Such
uncertainties in temporal labelling are currently ignored in model
training, resulting in learning mis-matched video-text correlation
with poor generalisation in test.
In this work,
we solve this problem by introducing
\method{full} (\method{abbr})
to accommodate flexible and adaptive activity temporal boundaries
towards modelling universally interpretable video-text correlation
with tolerance to underlying temporal uncertainties in pre-fixed annotations.
Specifically,
we construct elastic boundaries adaptively 
by mining and discovering
frame-wise temporal endpoints
that can
maximise the alignment between video segments and query sentences.
To enable both more accurate matching (segment content attention)
and more robust localisation (segment elastic boundaries),
we optimise the selection of frame-wise endpoints
subject to segment-wise contents
by a novel Guided Attention mechanism.
Extensive experiments on 
three video activity localisation
benchmarks
demonstrate compellingly the \method{abbr}'s advantages
over existing methods without modelling uncertainty.

\end{abstract}

\section{Introduction}
\label{sec:intro}

\begin{figure}
\centering
\includegraphics[width=1.\linewidth]{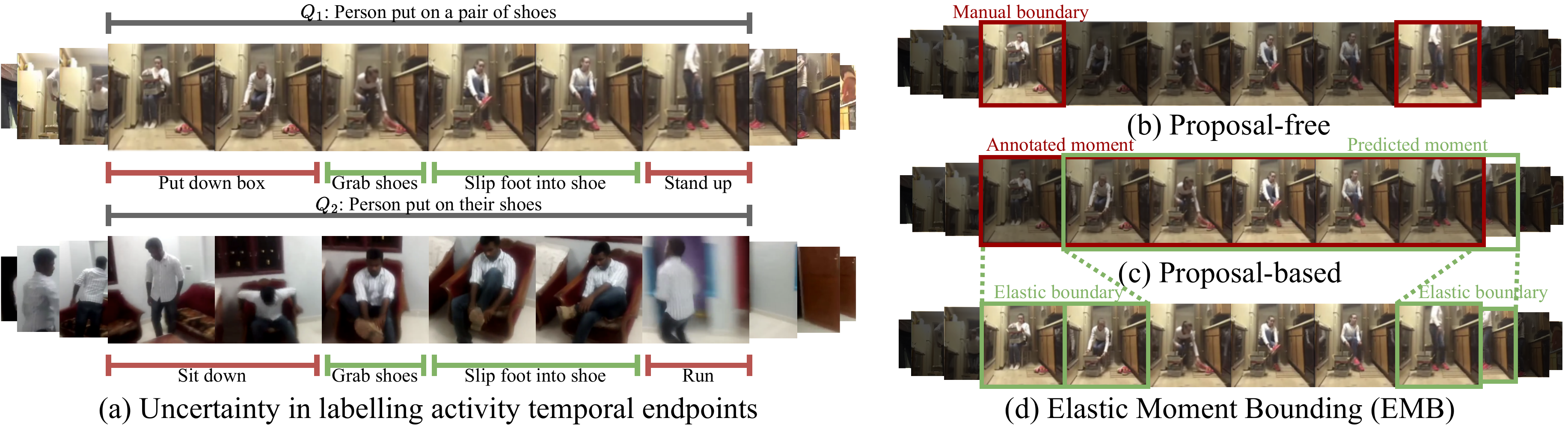}
\caption{
An illustration of different activity localisation methods.
(a) Activity's temporal boundaries are intrinsically uncertain
in manual labelling (break-down and highlighted in `red').
(b) Proposal-free methods learn to identify the frame-wise temporal endpoints.
(c) Proposal-based methods learn the holistic alignment
of video segments and query sentences between their feature spaces.
(d) \textit{\protect\method{full}} (\protect\method{abbr})
optimises simultaneously endpoints selection with maximisation of
segment content agreement between visual and textual representations.
}
\label{fig:ideas}
\end{figure}

The goal of video activity localisation is
to locate temporally video moments-of-interest (MoIs) of
a specific activity described by a natural language query of an
untrimmed continuous long video 
(often unscripted and unstructured) 
that contains many different activities~\cite{zhang2020vslnet,zhang2020tan2d,wang2020dpin}. 

One straightforward solution to the task,
denoted as proposal-free methods (Fig.~\ref{fig:ideas}~(b)),
is to predict directly the start and end frames of a target moment
that align to the given query~\cite{zeng2020drn,ghosh2019excl,zhang2020vslnet,nan2021ivg}.
Such paradigm deploys directly the fixed manual
activity endpoints labels for model training,
implicitly assuming these labels are well-defined and
ignoring uncertainties in the labels. 
However, unlike labelling object spatial bounding-boxes, 
there is a considerable variation in how activities
occur in unconstrained scenarios. 
There may not even be a precise definition of 
the exact temporal extent of an activity.
Fitting such uncertain temporal endpoints
will inevitably lead to semantically
mis-matched visual-textual correlations
which are not universally interpretable
and result in poor generalisation in test. 
For example, the two queries $Q_1$ and
$Q_2$ in Fig.~\ref{fig:ideas} (a) 
are semantically similar in describing `putting on shoes'. 
Nonetheless, the annotated activity (gray bars on top) 
for $Q_1$ starts from
putting down a box while $Q_2$ begins with sitting down on a sofa. 
By training a model with such uncertain temporal endpoints (Fig.~\ref{fig:ideas}~(b)),
the model is trained to match notably different visual features of
`putting down a box' and `sitting down on a sofa' with 
the same query on `putting on shoes'. 
Clearly, the model suffers from poor learning
due to uncertainty in visual cues. 
Moreover, as observed in~\cite{otani2020challengesmr}, 
\textit{the annotation bias can be
inconsistent from different annotators}. 
Giving the same videos
and query sentences to 5 different annotators, only $42\%$ and $35\%$
of their annotated activity boundary 
are mutually agreed 
(with at least $50\%$ IoU) 
on Charades-STA~\cite{gao2017charades} and
ActivityNet-Captions~\cite{krishna2017anet}, respectively. 
This highlights the extent of activity label uncertainties in
model training inherent to the current proposal-free methods, 
and the potential significant misinformation in training such models.
Another solution (Fig.~\ref{fig:ideas}~(c)) is to
generate many candidate proposals for a target moment
and aligns segment-level video features
with the query sentences~\cite{anne2017localizing,gao2017charades,zhang2020tan2d}.
By formulating the localisation task
as a matching problem,
the proposal-based methods
consider alignment by the whole moment with less focus on the
exact boundary matching~\cite{wang2021smin,wang2020dpin}. 
By doing so, it
can be less sensitive to the boundary labels but more reliance on
salient content (attention) between proposals and the target segment.
This can make them
more tolerant to the uncertainties in temporal annotations.
However, the problem of detecting accurately the start- and end-point of a
target activity moment \jh{remains unsolved
especially when learning without exhaustive moment proposals 
for efficiency concerns}.
 
In this work,
we introduce \textit{\method{full}} (\method{abbr})
to address the limitation of proposal-free paradigm
by modelling explicitly the label uncertainty in 
the temporal boundaries of an activity moment.
Instead of forcing a model to fit manually labelled \textit{rigid} activity endpoints, 
each MoIs are modelled by an elastic boundary
with a set of candidate endpoints. 
The model then learns to select
optimally from consistent visual-textual correlations 
among semantically similar activities.
This introduces model robustness to label uncertainty. 
Specifically, we conduct a
proposal-based segment-wise content alignment
in addition to learning of frame-wise boundary identification. 
As the predicted segment is required 
to be highly aligned with the query textual description, 
we represent the gap between the predicted endpoints and the
manual labelled endpoints as an elastic boundary
(Fig.~\ref{fig:ideas}~(d)). 
\jh{This process imposes explictly label uncertainties to model training.}
To enable activity localisation to be both more attention driven
(accurate) and sensitive to an elastic boundary (robust), 
we introduce an interaction between the segment-wise content representations
and frame-wise boundary features
by assembling representations through a \textit{Guided Attention} mechanism. 
The segment-wise boundary-guided attention 
helps minimise redundant frames 
in each elastic boundary 
whilst 
the frame-wise content-guided attention
highlighting transitional frames with
apparent visual changes indicating the potential start and end points of an activity.

We make three contributions in this work:
(1) We introduce a model to explore collaboratively both proposal-free and
proposal-based mechanisms for learning to 
detect more accurate activity temporal boundary localisation 
\textit{when training labels are inherently uncertain}.
We formulate a new \textit{\method{full}} (\method{abbr})
method to expand a manually annotated single pair of fixed activity endpoints 
to an elastic set.
(2) To reinforce directly robust content matching 
(the spirit of proposal-based) 
as a condition to accurate
endpoints localisation 
(the spirit of proposal-free) of activities in videos, 
we introduce a Guided Attention mechanism to explicitly
optimise frame-wise boundary visual features subject to
segment-wise content representations and vice versa, 
so to minimise redundant frames in each elastic boundary 
whilst highlighting frames signalling activity transitions subject to
segment content holistically. 
\jh{
(3) Our \method{abbr} model provides a state-of-the-art performance on
three video activity localisation benchmark datasets,
improving existing models 
that suffer from sensitivity to 
uncertainties in activity training labels.}


\section{Related work}
\label{sec:relate}

\keypoint{Proposal-based content alignment.}
By aggregating all the frames within a video segment
and aligning them holistically with the query sentences,
the segment-wise approaches~\cite{anne2017localizing,gao2017charades,zhang2020tan2d,zhang2019exploiting,ge2019mac}
are insensitive to the boundary
as its most salient
and semantically aligned parts
are not necessarily at its two ends.
The endpoint frames play a significant role
to help differentiate video moments from their overlapping counterparts
containing redundant frames,
hence,
critical for video activity localisation.
Therefore, we explicitly associate 
the segment-wise content information
with the frame-wise boundary information
and complement them by each other through 
a novel guided attention mechanism.

\keypoint{Proposal-free boundary identification.}
In contrast,
the proposal-free methods
learn to directly regress the start and end timestamps of the target moments
or predict the per-frame probabilities of being the endpoints%
~\cite{chen2018temporally,zeng2020drn,ghosh2019excl,zhang2020vslnet,nan2021ivg,zhou2021embracing,zhao2021cpn,li2021cpnet}.
In either case,
they take the temporal boundaries provided manually
as the oracles 
for learning exactly the same predictions.
However,
this is prone to be misled by
the uncertainty in manual labels
and results in less generalisable models.
To cope with that,
we train our \method{abbr} model
to identify the target boundary from
reliable candidate start and end spans (sets of frames)
rather than fitting the single pre-fixed manual endpoints,
so as to derive consistent video-text correlations
from semantically similar activities
that are universally interpretable.

\keypoint{Joint content-boundary learning.}
There are a few recent attempts~\cite{wang2021smin,wang2020dpin,xiao2021bpnet}
on locating video activity
jointly by the proposal-based and proposal-free strategies.
They mostly explored the interaction of 
frame's and segment's feature representations
for better video comprehension.
In this work,
we study the combination of the two strategies 
for attention learning of activity temporal boundary conditions
beyond feature learning for activity representation.
We augment the fixed manual labels
by the video segments selected according to 
their content alignments with query sentences
to help improve the robustness of 
temporal endpoints identification when there is boundary uncertainty.

\keypoint{Temporal boundary uncertainty.}
Recently,
Otani \etal~\cite{otani2020challengesmr} quantitatively studied 
the label uncertainty problems on video activity localisation
by collecting multiple boundaries for the same activities 
from different annotators,
the results highlighted the extent of uncertainty 
in the temporal annotations.
However, Otani \etal~\cite{otani2020challengesmr}
did not explicitly propose a solution to the problem.
DeNet~\cite{zhou2021embracing}, 
on the other hand,
addressed it \wrt\ the variety of language descriptions,
\ie, the same video activity can be described semantically in different ways.
They generated different copies of the same query sentences by 
perturbing the ``modified'' phrases (adjective, adverb and \etc)
so to predict diverse boundaries for the same video activities.
Rather than studying the uncertainty
from the perspective of semantic description,
we analyse 
the uncertainties in activity temporal boundary annotations, 
which is intrinsically harder to avoid.


\section{Learning Localisation with Uncertainty}
\label{sec:method}
\begin{figure}
\centering
\includegraphics[width=1.0\linewidth]{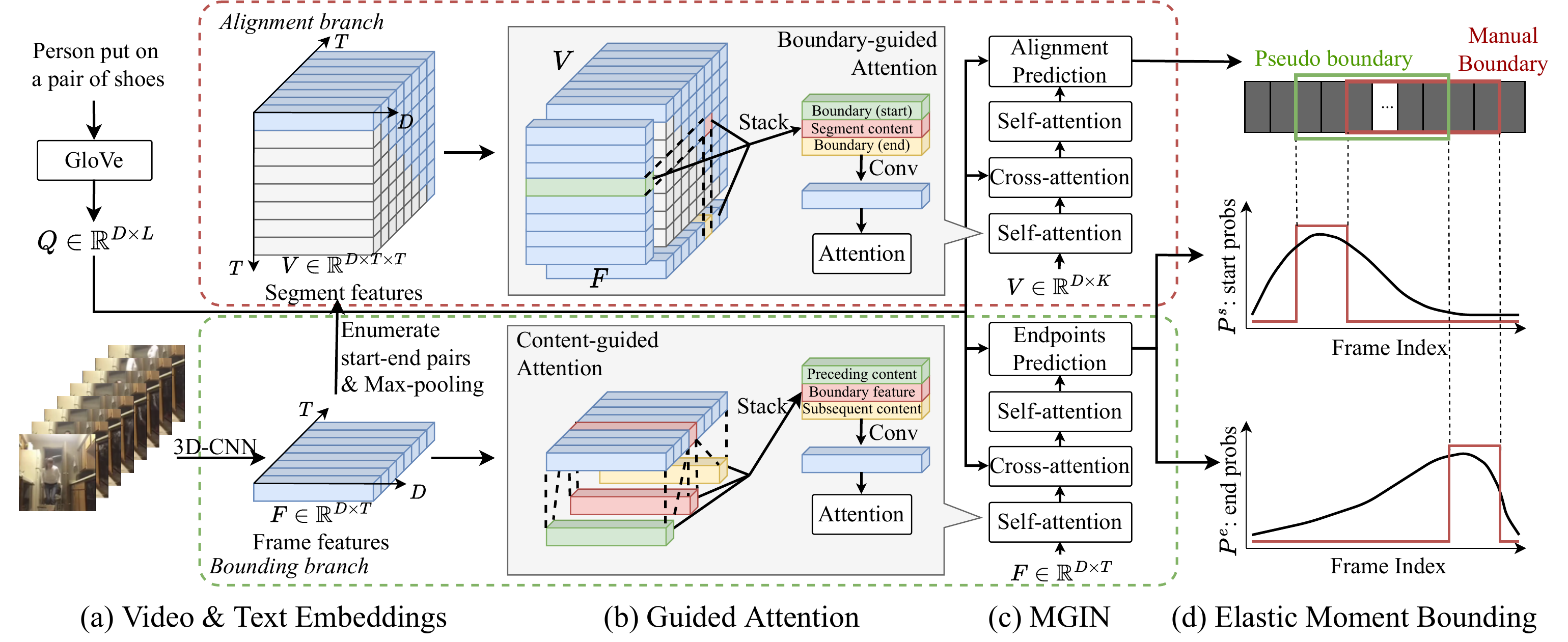}
\caption{
An overview of \textit{\protect\method{full}} (\protect\method{abbr}). 
\textbf{(a)} 
The \protect\method{abbr} model 
takes the pretrained 3D-CNN features and the GloVe embeddings
as inputs.
It consists of an `Alignment branch' (red dashed box)
learning the semantic content alignment of video segments and query sentences,
and a `Bounding branch' (green dashed box) 
to predict the temporal endpoints of target activity moments.
Both branches are subject to 
\textbf{(b)} a Guided Attention mechansim with
both self- and cross-modalities attention in
\textbf{(c)} a Multi-grained Interaction Network (MGIN).
\textbf{(d)}
The most confidently aligned video segments
predicted by the alignment branch
are then selected for
constructing the elastic boundary
to optimise the endpoints predictions from the bounding branch. 
}
\label{fig:overview}
\end{figure}
Given the feature representations of an untrimmed videos 
$\bm{F}$ composed of $T$ frames,
and that of a natural language sentence 
$\bm{Q}$ of $L$ words,
the objective of video activity localisation is to
identify the temporal boundary of a target moment $(S, E)$
-- activity endpoints --
so that the video segment $\{\bm{f}_t\}_{t={S}}^{E}$
matches with $\bm{Q}$ in semantics.
It is challenging
to acquire high-level semantic understandings of either videos or sentences,
let alone aligning them to precisely 
locate the temporal endpoints of a specific activity instance.

In this work,
we study the problem of model learning subject to temporal label uncertainty
which is inherent to manual video annotation and more importantly
not shared in unseen new test videos or language descriptions.
To that end,
we propose
an \textit{\method{full}} (\method{abbr}) model (Fig.~\ref{fig:overview}).
The \method{abbr} model first 
predicts the per-frame probabilities
to be the temporal endpoints of a target moment
(Fig.~\ref{fig:overview}'s green dashed box) 
by a \textit{Multi-grained Interaction Network} (MGIN)
(Fig.~\ref{fig:overview}~(c))
incorporating with
a \textit{Guided Attention} mechanism (Fig.~\ref{fig:overview}~(b)).
\method{abbr} then optimises the frame-wise probabilities
by mining multiple candidate endpoints beyond the manual annotated ones.
The candidate endpoints are discovered by
an auxiliary alignment branch (Fig.~\ref{fig:overview}'s red dashed box).
The alignment branch explores 
the visual-textual content aligment at segment-level,
which is less sensitive to exact endpoints annotations
so more robust to uncertainty.
By doing so,
we construct an \textit{elastic boundary}
interpretable 
universally for semantically similar activities with endpoints uncertainty.

\subsection{Temporal Endpoints Identification}
Our elastic moment bounding
is a generic formulation deployable in any multi-modal backbone deep networks.
Here, we start with the VSLNet~\cite{zhang2020vslnet}
and reconstruct it by introducing a Guided Attention mechanism
to form a multi-grained interaction network (MGIN). 
The overall pipeline of MGIN is shown in Fig.~\ref{fig:overview} (c)
to first encode the video $\bm{F}$ and the sentence $\bm{Q}$
by attention both within (self) and across modalities,
then predict the frame-wise endpoint probabilities
by the joint-modal representations 
fused by context-query attention~\cite{seo2016bidirectional,xiong2016dynamic,yu2018fast}.

Adopting the convention~\cite{zhang2020vslnet,nan2021ivg,wang2021smin},
we represent video frames
by a pre-trained 3D-CNN model~\cite{carreira2017i3d}
as $\bm{F}=\{\bm{f}_t\}_{t=1}^T \in \mathbb{R}^{D^v\times T}$
and the query sentence by 
the GloVe embeddings~\cite{pennington2014glove} of words 
$\bm{Q}=\{\bm{w}_l\}_{l=1}^L \in \mathbb{R}^{D^q\times L}$.
To facilitate cross-modal feature interactions,
we map both the representations to have the same dimension $D$
by two independent linear projections,
\ie, $\bm{F} \leftarrow \fc(\bm{F}) \in \mathbb{R}^{D\times T}$ and 
$\bm{Q} \leftarrow \fc(\bm{Q}) \in \mathbb{R}^{D\times L}$.

\keypoint{Vision-language attention representation.}
We deploy attentive encoding~\cite{vaswani2017attention,huang2021cross}
for both the visual and textual representations
to explore the dependencies among elements in both.
In general,
to encode a target sequence $\bm{X}^t\in\mathbb{R}^{D\times L^t}$ of $L^t$ elements
with the help of a reference sequence 
$\bm{X}^r\in\mathbb{R}^{D\times L^r}$ in size $L^r$,
we first compute an attention matrix $\bm{A}$ 
indicating the pairwise target-reference correlations,
then represent each target element by its correlated references:
\begin{gather}
\bm{A} = \fc(\bm{X}^t)^\top\fc(\bm{X}^r)/\sqrt{D} \in \mathbb{R}^{L^t\times L^r} \label{eq:attention} \\
g(\bm{X}^t, \bm{X}^r) = \bm{X}^t + \fc(\bm{X}^r){\softmax(\bm{A})}^\top \in \mathbb{R}^{D\times L^t}.
\label{eq:attention_encode}
\end{gather}
An attention layer formulated in Eq.~\eqref{eq:attention_encode}
is parameterised by three independent fully-connected layers.
Our MGIN shown in Fig.~\ref{fig:overview} (c)
is constructed by both self-attention within modalities:
$\bm{F}\leftarrow g(\bm{F}, \bm{F})$, $\bm{Q}\leftarrow g(\bm{Q}, \bm{Q})$
for context exploration
and cross-attention between modalities:
$\bm{F}\leftarrow g(\bm{F}, \bm{Q})$, $\bm{Q}\leftarrow g(\bm{Q}, \bm{F})$
to learn the semantic correlations between video frames and query words.

\keypoint{Guided attention.}
To effectively locate the temporal endpoints of activities,
it is essential for the model to be aware of
not only what is shown in each individual frame
but also what's different before and after it.
As a simple example,
the starting point of an activity `person puts on shoes'
should not be arbitrary frames involving shoes-like objects
in-between the period
but be consistent 
like when the shoes first appear to interact with the person.
Therefore,
we propose a content-guided attention module 
(Fig.~\ref{fig:overview} (b)'s bottom)
to explicitly encode the preceding and subsequent content information
of each frame into its representation:
\begin{equation}
\begin{gathered}
\bm{F}_\text{pre} = \{\agg(\{\bm{f}_i\}_{i=1}^{t})\}_{t=1}^T \in \mathbb{R}^{D\times T}, \\
\bm{F}_\text{sub} = \{\agg(\{\bm{f}_i\}_{i=t}^T)\}_{t=1}^T \in \mathbb{R}^{D\times T},\\
\tilde{\bm{F}} = \text{Conv2d}(\{\bm{F}, \bm{F}_\text{pre}, \bm{F}_\text{sub}\}) \in \mathbb{R}^{D\times T}.
\label{eq:guided_frame}
\end{gathered}
\end{equation}
The feature $\agg(\{\bm{f}_i\}_{i=1}^t) \in \mathbb{R}^D$ in Eq.~\ref{eq:guided_frame}
aggregate all the frames before $\bm{f}_t$
by max-pooling as its \underline{pre}ceding content representation.
Similarly,
the \underline{sub}sequent content of the $t$-th frame
is obtained by 
$\agg(\{\bm{f}_i\}_{i=t}^T)$.
Both the preceding $\bm{F}_\text{pre}$ and subsequent $\bm{F}_\text{sub}$ content features 
are then stacked and assembled 
with the frame-wise representations $\bm{F}$
by a 2D convolution layer.
After that,
the content-guided representations of video frames $\tilde{\bm{F}}$
are used for attentive encoding (Eq.~\eqref{eq:attention_encode})
both within
$\bm{F} \leftarrow g(\tilde{\bm{F}}, \tilde{\bm{F}})$
and across modalities
$\bm{F} \leftarrow g(\tilde{\bm{F}}, \bm{Q})$.

\keypoint{Boundary prediction.}
Given a video $\bm{F}\in\mathbb{R}^{D\times T}$
and sentence $\bm{Q}\in\mathbb{R}^{D\times L}$ representations,
we estimate the frame-wise endpoint probabilities by computing 
context-query attention~\cite{seo2016bidirectional,xiong2016dynamic,yu2018fast}, 
same as the baseline~\cite{zhang2020vslnet}. 
It is defined as
\begin{equation}
\begin{gathered}
(\bm{p}^s, \bm{p}^e) \!=\!
  \softmax(\text{LSTM}(\hat{\bm{F}}\odot\bm{h})), \textrm{where} \,\,
\bm{h} \!=\! \sigma(\text{Conv1d}(\hat{\bm{F}}\Vert \bm{q})) \in
\mathbb{R}^{1\times T}, \\
\hat{\bm{F}} = H(\bm{F}, \bm{Q}) = 
\fc(\bm{F}\Vert\bm{X}^{v2q}\Vert\bm{F}\odot\bm{X}^{v2q}\Vert\bm{F}\odot\bm{X}^{q2v})
\in \mathbb{R}^{D\times T}; \,\, \textrm{and} \\ 
\bm{A} = \frac{{\fc(\bm{F})}^\top\fc(\bm{Q})}{\sqrt{D}},\ 
\bm{X}^{v2q} = \bm{Q}{\bm{A}^r}^\top,\ 
\bm{X}^{q2v} = \bm{F}\bm{A}^r{\bm{A}^c}^\top.
\label{eq:condition_predict}
\end{gathered}
\end{equation}
In Eq.~\eqref{eq:condition_predict}, 
we predict the frame-wise endpoint probabilities
by two stacked LSTM.
This is based on fusing the two modalities $\bm{F}$ and $\bm{Q}$ by function $H$
then rescale the per-frame fused feature $\hat{\bm{F}} \in \mathbb{R}^{D\times T}$
using their estimated likelihood $\bm{h} \in \mathbb{R}^{1\times T}$
of being foreground 
to suppress any distractions from redundant frames.
Matrix $\bm{A} \in \mathbb{R}^{T\times L}$ 
consists of frame-to-word correlation scores;
$\bm{A}^r$ and $\bm{A}^c$ are its row and column-wise softmax normalised copies.
The $\bm{q}$ are the sentence-level representations
from weighted sum of words~\cite{bahdanau2014neural};
$(\cdot\Vert\cdot)$ stands for concatenation
(broadcast if necessary)
while $\odot$ is the Hadamard Product.

\subsection{Elastic Moment Bounding}
Given the uncertainty and ambiguity in manually annotated activity temporal boundaries, 
it is ineffective to decide heuristically and universally
which frames and how many of them 
should be taken as the candidate endpoints $(\tilde{\bm{S}}, \tilde{\bm{E}})$
for different video activities.
To address this problem, 
we formulate an auxiliary alignment branch in the model
to learn the visual-textual content mapping per each video segment.
It serves as an additional self-learning ``annotator''
to expand the given single pair of manually annotated boundaries
into candidate endpoints proposal sets
tailored for individual activities.

\keypoint{Elastic boundary construction.}
As shown in Fig.~\ref{fig:overview}'s red dashed box,
we first generate a 2D feature map~\cite{zhang2020tan2d} 
by enumerating pairwise start-end frames
to represent $K=T\times T$ video segments $\bm{V}=\{\bm{v}_k\}_{k=1}^K\in\mathbb{R}^{D\times K}$
as the proposals for a target moment.
We flatten the 2D map here for clarity.
The $k$-th proposal with the temporal boundary of $(t^s_k, t^e_k)$ 
is represented by max-pooling the frames it is composed of
$\bm{v}_k = \agg(\{\bm{f}_t \vert \forall t \in [t^s_k, t^e_k]\})$.
The segment-wise representations will then be fed into 
an independent MGIN equipped with 
\textit{boundary-guided} attention modules (Fig.~\ref{fig:overview} (b)'s top)
for visual encoding.
Similar as in the \textit{content-guided} attention for video frames,
we explicitly assemble the frame-wise boundary features 
with the content representations of video segments
to encourage boundary-sensitive content alignment:
\begin{equation}
\begin{gathered}
\bm{V}_\text{sta} = \{\bm{f}_{{t}^s_k}\}_{k=1}^K \in \mathbb{R}^{D\times K},\ 
\bm{V}_\text{end} = \{\bm{f}_{{t}^e_k}\}_{k=1}^K \in \mathbb{R}^{D\times K}, \\
\tilde{\bm{V}} = \text{Conv2d}(\{\bm{V}, \bm{V}_\text{sta}, \bm{V}_\text{end}\}) \in \mathbb{R}^{D\times K}.
\label{eq:guided_segment}
\end{gathered}
\end{equation}
The features $\bm{V}_\text{sta}$ and $\bm{V}_\text{end}$ in Eq.~\eqref{eq:guided_segment}
are the representations of the \underline{sta}rt and \underline{end} frames 
for each of the $K$ proposals.
They are stacked and assembled with the segment-wise content features $\bm{V}$
to derive the boundary guided segment representations $\tilde{\bm{V}}$
by a 2D convolution layer.
Such boundary-guided attention share a similar spirit
with temporal pyramid pooling~\cite{zhao2017temporal}, 
that is
to explicitly encode the temporal structure 
into segment's representation
so to be sensitive to its boundary.
$\tilde{\bm{V}}$ is then used for attentive encoding (Eq.~\eqref{eq:attention_encode})
within $\bm{V} \leftarrow g(\tilde{\bm{V}}, \tilde{\bm{V}})$
and across $\bm{V} \leftarrow g(\tilde{\bm{V}}, \bm{Q})$
modalities.

Given the segment-level video representations $\bm{V}$,
we fuse them with the sentence features
by $H$ defined in Eq.~\eqref{eq:condition_predict},
and re-arrange it to be a 2D feature map
then predict the per-proposal alignment scores by a 2D convolution layer:
\begin{equation}
\bm{p}^a = \sigma(\text{Conv2d}(H(\bm{V}, \bm{Q})))\ s.t.\ p^a_k \in (0, 1)\ \forall k \in [1, K].
\label{eq:conv_predict}
\end{equation}
The segment-wise alignment scores $\bm{p}^a$
activated by the Sigmoid function $\sigma$
is then supervised by
the temporal overlaps between every proposals and the manual boundary:
\begin{equation}
\begin{gathered}
\alpha_k = \text{IoU}(({t}^s_k, {t}^e_k), (S, E))\\
 y^a_k=\left\{
\begin{array}{ll}
1, & \text{if}\ \alpha_k \geq \tau_u \\
0, & \text{if}\ \alpha_k < \tau_l \\
\alpha_k, & \text{otherwise}
\end{array}\right. \\
\mathcal{L}_\text{align}(\bm{V},\bm{Q}, S,E) = \bce(\bm{y}^a, \bm{p}^a).
\end{gathered}
\label{eq:loss_align}
\end{equation}
The notations $\tau_u$ and $\tau_l$ are the upper and lower overlap thresholds
to control the flexibility of video-text alignment, 
which are set to $0.7$ and $0.3$ respectively
as in~\cite{zhang2020tan2d}.
With the learned segment-wise alignment scores $\bm{p}^a$,
we take the boundary $({t}^s_{k^*}, {t}^e_{k^*})$ of 
the most confident proposal
with the greatest predicted score $p^a_{k*} \geq p^a_{k}\ \forall k \in [1, K]$
as the pseudo boundary 
and construct the corresponding candidate endpoint sets by:
\begin{equation}
\begin{gathered}
\tilde{\bm{S}} = [\min({t}^s_{k^*}, S), \max({t}^s_{k^*}, S)],\quad
\tilde{\bm{E}} = [\min({t}^e_{k^*}, E), \max({t}^e_{k^*}, E)].
\end{gathered}
\label{eq:candidate}
\end{equation}
We customise the candidate endpoint sets
for every individual activity
by exploring the content alignments
between video segments and query sentences,
\ie, elastic boundary.
This is intuitively more reliable than
applying label smoothing globally~\cite{wang2021smin,xiao2021bpnet}
without considering video context and language semantics.

\keypoint{Reliability \vs\ flexibility.}
Introducing too many candidate endpoints
that are semantically irrelevant to the query sentences
is prone to distracting the model from learning effective visual-textual correlations,
especially at the early stage of training
a randomly initialised model
which is likely to yield inaccurate pseudo boundaries
$({t}^s_{k^*}, {t}^e_{k^*})$.
Therefore,
we balance the reliability and flexibility
of our elastic boundary
by a controllable threshold $\tau$:
\begin{equation}
{k*} = \arg\max_{k} \bm{p}^a\quad s.t.\quad \alpha_k \geq \tau.
\label{eq:pseudo}
\end{equation}
The $\alpha_k$ in Eq.~\eqref{eq:pseudo}
implies the overlap between the $k$-th proposal
and the manual boundary,
whilst
the threshold $\tau$
serving as a tradeoff between flexibility and reliability
so that only the sufficiently overlapped proposals 
will be selected for constructing
the elastic boundary in Eq.~\eqref{eq:candidate}.

\keypoint{Learning from elastic boundary.}
With the elastic boundary $(\tilde{\bm{S}}, \tilde{\bm{E}})$,
we formulate the boundary supervision signals
to maximise the sum of the candidate endpoint's probabilities
obtained in Eq.~\eqref{eq:condition_predict}:
\begin{equation}
\mathcal{L}_\text{bound}(\bm{F},\bm{Q}, S, E) = 
-\log(\sum_{t\in\tilde{\bm{S}}}p^s_t) -\log(\sum_{t\in\tilde{\bm{E}}}p^e_t).\\
\label{eq:loss_bound}
\end{equation}
Comparing with the commonly adopted frame-wise supervision
which trains $\bm{p}^s$ and $\bm{p}^e$ to be one-hot~\cite{zhang2020vslnet,nan2021ivg},
we provide in Eq.~\eqref{eq:loss_bound} a more flexible boundary to the target moments
so that the model can learn in a data-driven manner
to select the endpoints
beyond the manual boundary
and ignore the unconcerned actions involved.

\subsection{Model Training and Inference}

\keypoint{Inference.}
We consider two scenarios when predicting the boundary of video activity:
\textbf{(a) DET:} following the standard protocol of the task~\cite{gao2017charades,heilbron2015anet},
we predict a determined boundary enclosed by a single start and end frames
according to the outputs of bounding branch in a maximum likelihood manner
\begin{equation}
\hat{S} = \arg\max_{t}\bm{p}^s,\quad
\hat{E} = \arg\max_{t}\bm{p}^e,
\label{eq:infer}
\end{equation}
where $\hat{S}$ and $\hat{E}$ are the predicted 
start and end frame indices of a video that are corresponding to a given query.
\textbf{(b) ELA:}
considering the uncertain nature of temporal boundary,
it is more intuitive to estimate the endpoints of video activity
by temporal spans rather than specific frames.
Our model is able to predict also an elastic boundary
in a similar way as in training:
\begin{equation}
\begin{gathered}
\hat{\bm{S}} = [\min({t}^s_{k^*}, \hat{S}), \max({t}^s_{k^*}, \hat{S})],\quad
\hat{\bm{E}} = [\min({t}^e_{k^*}, \hat{E}), \max({t}^e_{k^*}, \hat{E})].
\end{gathered}
\label{eq:infer_elastic}
\end{equation}
In Eq.~\eqref{eq:infer_elastic}, we denote $\hat{\bm{S}}$ and $\hat{\bm{E}}$ in bold
to indicate a set of candidate endpoints,
and differentiate them from the determined boundary in Eq.~\eqref{eq:infer}.
The $(t^s_{k*}, t^e_{k*})$ is the boundary of the most confident proposals
selected from the alignment branch 
without constraint on their overlaps to the ground-truth (Eq.~\eqref{eq:pseudo}).

\keypoint{Training.} 
In addition to $\mathcal{L}_\text{bound}$ and $\mathcal{L}_\text{align}$,
we follow the baseline to learn $\bm{h}$ in Eq.~\eqref{eq:condition_predict}
by a binary cross-entropy loss to 
highlight foreground video content:
\begin{equation}
\mathcal{L}_\text{high}(\bm{F}, \bm{Q}, S, E) = \bce(\bm{y}^h, \bm{h}),\quad
y^h_t = \mathbbm{1}[\min(\tilde{\bm{S}}) \leq t \leq \max(\tilde{\bm{E}})].
\label{eq:loss_sup}
\end{equation}
Note that,
the boundary $(\min(\tilde{\bm{S}}), \max(\tilde{\bm{E}}))$ is 
also extended as in \cite{zhang2020vslnet} to
encourage the model to be sensitive to subtle visual changes 
around the temporal endpoints.
The overall loss function of \method{abbr} is then formulated as:
\begin{equation}
\mathcal{L} = 
\lambda_1 \mathcal{L}_\text{bound} +
\lambda_2 \mathcal{L}_\text{align} +
\lambda_3 \mathcal{L}_\text{high}
\label{eq:loss}
\end{equation}
The \method{abbr} model is optimised end-to-end by
stochastic gradient descent.
Its overall training process is summarised in Alg.~\ref{alg:emb}.

\begin{algorithm}[ht]
    \caption{\protect\method{full}} 
    \label{alg:emb}
    \textbf{Input:} 
    An untrimmed video $\bm{F}$,
    a query sentence $\bm{Q}$,
    a temporal boundary $(S, E)$. \\
    \textbf{Output:}
    An updated video activity localisation model. \\
    Encode frames by content-guided attention via Eq.~\eqref{eq:attention_encode}\eqref{eq:guided_frame}; \\
    Fuse frames with query and predict per-frame endpoint probabilities via Eq.~\eqref{eq:condition_predict}; \\
    Construct 2D feature map of proposals; \\
    Encode proposals by boundary-guided attention via Eq.~\eqref{eq:attention_encode}\eqref{eq:guided_segment}; \\
    Fuse proposals with query and predict proposal-query alignment scores via Eq.~\eqref{eq:conv_predict}; \\
    Construct the elastic boundary via Eq.~\eqref{eq:candidate}\eqref{eq:pseudo}; \\
    Optimise model weights by minimising $\mathcal{L}$ via Eq.~\eqref{eq:loss}.
\end{algorithm}

\section{Experiments}
\label{sec:exp}

\keypoint{Datasets.}
We evaluated the proposed \method{abbr} model
on three widely adopted video activity localisation
benchmark datasets:
\textbf{(1)} TACoS~\cite{regneri2013tacos,rohrbach2012script},
\textbf{(3)} Charades-STA~\cite{gao2017charades,sigurdsson2016charades} and
\textbf{(2)} ActivityNet-Captions~\cite{krishna2017anet,heilbron2015anet}.
Their different data characteristics are summarised in Table~\ref{tab:data}.
Among the three datasets,
the raw videos in TACoS have the longest durations 
while that of its MoIs are shortest in contrast, 
which means that the video activities are temporally covering
less than $2\%$ of the complete videos on average.
Therefore,
the videos in TACoS contain a lot of redundancy in terms of every MoIs.
On the other hand,
the ActivityNet-Captions is very different from TACoS
whose video activities temporally cover 
much larger proportions of the videos ($\sim30\%$) than the other two.

\keypoint{Performance Metrics.}
We followed the common practices~\cite{zhang2020vslnet,wang2021smin,nan2021ivg} 
to measure the quality of our video activity localisation results
by their average recall rate
at different temporal IoU thresholds (IoU@$m$).
The predicted boundary $(\hat{S}, \hat{E})$
of a MoI
is considered correct
if its IoU with the manual temporal label $(S, E)$
is greater than the thresholds $m$
which are predefined as $m=\{0.3, 0.5, 0.7\}$.
Besides,
we also reported the mean IoU (mIoU)
of all predictions with their corresponding ground-truth
to show the average overlaps between
the predicted and manual boundaries.
For our elastic boundary,
we enumerate all the start-end pairs from
$\hat{\bm{S}}$ and $\hat{\bm{E}}$ (Eq.~\eqref{eq:infer_elastic}) respectively.
If a manual boundary's overlap to any of the combinations
is greater than the IoU threshold,
we consider it is correctly predicted.
\begin{table}
\setlength{\tabcolsep}{0.35cm}
\begin{center}
\caption{Statistics of datasets.
$L^v$ and $L^m$ are the average lengths of videos and MoIs, respectively.
$L^q$ is the average number of words in query sentences. 
}
\label{tab:data}
\begin{tabular}{l|c|c|c|c|c|c}
\noalign{\smallskip}\hline
Datset & \#Train & \#Val &\#Test & $L^v$ & $L^q$ & $L^m$ \\ \hline
TACoS~\cite{regneri2013tacos} & 10,146 & 4,589 & 4,083 & 287.14s & 10.1 & 5.45s \\ \hline
ANet~\cite{krishna2017anet} & 37,421 & 17,031 & 17,505 & 117.61s & 14.8 & 36.18s \\ \hline
Charades~\cite{gao2017charades} & 12,408 & - & 3,720 & 30.59s & 7.2 & 8.22s \\ \hline
\end{tabular}
\end{center}
\end{table}

\keypoint{Implementation Details.}
We adopted the video features provided by
our baseline model~\cite{zhang2020vslnet}
and the 300D GloVe~\cite{pennington2014glove} embeddings
to encode the 
video and text inputs, respectively. 
We downsampled videos to have $128$ frames at most by max-pooling
and zero-padded the shorter ones.
The outputs of all the hidden layers
were 128D as in~\cite{zhang2020vslnet}
and the multi-head variant~\cite{vaswani2017attention}
of the attention layer in Eq.~\eqref{eq:attention}
was used with $8$ heads followed by
layer normalisation and random dropout at $0.2$.
Cosine positional embeddings were applied to the inputs.
The \method{abbr} model was trained
for 100 epochs with a batch size of $16$.
It was optimised by an Adam optimiser 
using a linearly decaying learning rate of $5e-4$
and gradient clipping of $1.0$.
In the alignment branch,
we downsampled the videos to have $16$ clips
by the max-pooling of every $8$ continuous frames
for constructing the 2D feature maps of video segments
to avoid over-dense proposals.
The threshold $\tau$ in Eq.~\eqref{eq:pseudo}
was initiated to be $1$ and progressively decreased to $0.5$.
The weights of losses were empirically set to
$\lambda_1=\lambda_2=1$ and $ \lambda_3=5$
in all the datasets.

\subsection{Comparisons to the State-of-the-art}
As shown in Table~\ref{tab:sota},
the determined boundary yielded by \method{abbr} (DET)
outperforms the baseline VSLNet~\cite{zhang2020vslnet}
by non-negligble margins on all tests.
The more recent IVG~\cite{nan2021ivg} shares the same baseline as \method{abbr}.
The notable performance advantages of \method{abbr} over both of them
demonstrate its non-trivial improvements. 
Furthermore, 
\method{abbr} surpasses the state-of-the-art methods
on TACoS against all the performance metrics
while remaining its competitiveness on the other two datasets.
Among the three datasets, 
TACoS poses the hardest test with the
longest average untrimmed videos and 
the shortest activity moments 
(see Table~\ref{tab:data}). 
That is, TACoS exhibits more
realistic scenarios for activity localisation test. 
In this context, \method{abbr} shows its
advantage 
most clearly when the untrimmed videos
are longer whilst the video MoIs are sparse and far between.

Moreover, 
table~\ref{tab:sota} shows also the clear performance advantages
of the elastic boundary predicted by our \method{abbr} (ELA) model
over a wide range of the state-of-the-art methods.
When constructing the elastic boundaries in inference,
over 80\% of the predictions pairs 
yielded by the alignment and bounding branches 
are consistent with each other ($\text{IoU}>0.5$).
Therefore, the performance improvements we obtained 
is not due to over-dense sampling of 
the potential boundaries.
For fairer comparisons,
we took the adjacent frames before and after
the endpoints predicted by VSLNet
to generate multiple candidate boundaries for its evaluation.
The number of frames is set to be $10\%$ of the moment length
so that the density of candidate boundaries is consistent with ours.
Although clear performance gains are observed,
the improvements
from such a \textit{global} shifting strategy
are less competitive to our per-sample \textit{adaptive} designs
due to missing considerations of sample-dependent bias.
\begin{table}[ht]
\scriptsize
\setlength{\tabcolsep}{2.4pt}
\begin{center}
\caption{Performance comparisons to the state-of-the-art models
on three video activity localisation benchmark datasets.
The first and second best results are highlighted in red and blue, respectively. 
The `DET' modifier of EMB
stands for the determined boundary predicted in Eq.~\eqref{eq:infer}
while `ELA' is the elastic boundary (Eq.~\eqref{eq:infer_elastic}).
The symbol $\dagger$ denotes the reproduced results of our baseline model
under the strictly identical setups
using the code from authors
and $\star$ indicates multi-candidate predictions.
}
\label{tab:sota}
\begin{tabular}{l||c|c|c|c||c|c|c|c||c|c|c|c}
\noalign{\smallskip}\hline
\multirow{3}{*}{Method} & 
\multicolumn{4}{c||}{TACoS~\cite{regneri2013tacos}} & 
\multicolumn{4}{c||}{Charades-STA~\cite{gao2017charades}} &
\multicolumn{4}{c}{ActivityNet-Captions~\cite{krishna2017anet}} \\ \cline{2-13}
& \multirow{2}{*}{mIoU} & \multicolumn{3}{c||}{$\text{IoU}@m$}
& \multirow{2}{*}{mIoU} & \multicolumn{3}{c||}{$\text{IoU}@m$}
& \multirow{2}{*}{mIoU} & \multicolumn{3}{c}{$\text{IoU}@m$} \\ 
\cline{3-5}\cline{7-9}\cline{11-13}
&& 0.3 & 0.5 & 0.7 
&& 0.3 & 0.5 & 0.7 
&& 0.3 & 0.5 & 0.7 \\ \hline
\hline
VSLNet~\cite{zhang2020vslnet} 
& 24.11 & 29.61 & 24.27 & {20.03}
& 45.15 & 64.30 & 47.31 & 30.19
& 43.19 & 63.16 & 43.22 & 26.16 \\
IVG~\cite{nan2021ivg} 
& 28.26 & 38.84 & 29.07 & 19.05
& 48.02 & 67.63 & 50.24 & 32.88 
& {44.21} & {63.22} & 43.83 & 27.10 \\
2D-TAN~\cite{zhang2020tan2d} 
&   -   & 37.29 & 25.32 &   -  
&   -   &   -   & 39.70 & 23.31
&   -   & 59.45 & 44.51 & 26.54 \\
LGI~\cite{mun2020lgi} 
&   -   &   -   &   -   &   -  
& {51.38} & \best{72.96} & 59.46 & 35.48
& 41.13 & 58.52 & 41.51 & 23.07 \\
DPIN~\cite{wang2020dpin} 
&   -   & 46.74 & 32.92 &   -  
&   -   &   -   & 47.98 & 26.96
&   -   & 62.40 & {47.27} & {28.31} \\
DRN~\cite{zeng2020drn} 
&   -   &   -   & 23.17 &   -   
&   -   &   -   & 53.09 & 31.75
&   -   &   -   & 45.45 & 24.36 \\
SCDM~\cite{yuan2019scdm} 
&   -   & 26.11 & 21.17 &   -   
&   -   &   -   & 54.44 & 33.43
&   -   & 54.80 & 36.75 & 19.86 \\
BPNet~\cite{xiao2021bpnet} 
& 19.53 & 25.93 & 20.96 & 14.08
& 46.34 & 65.48 & 50.75 & 31.64
& 42.11 & 58.98 & 42.07 & 24.69 \\
CPNet~\cite{li2021cpnet} 
& {28.69} & 42.61 & 28.29 &   -  
& \scnd{52.00} &   -   & {60.27} & {38.74}
& 40.65 &   -   & 40.56 & 21.63 \\
CPN~\cite{zhao2021cpn} 
& \scnd{34.63} & \scnd{48.29} & \scnd{36.58} & \scnd{21.25}
& 51.85 & \scnd{72.94} & 56.70 & 36.62
& \best{45.70} & 62.81 & 45.10 & \scnd{28.10} \\
DeNet~\cite{zhou2021embracing} 
&   -   &   -   &   -   &   -  
&   -   &   -   & 59.75 & 38.52
&   -   & 61.93 & 43.79 &   -   \\
CBLN~\cite{zhao2021cpn} 
&   -   & 38.98 & 27.65 &   -  
&   -   &   -   & \scnd{61.13} & 38.22
&   -   & \best{66.34} & \scnd{48.12} & 27.60 \\
SMIN~\cite{wang2021smin} 
&   -   & {48.01} & {35.24} &   -  
&   -   &   -   & \best{64.06} & \best{40.75}
&   -   &   -   & \best{48.46} & \best{30.34} \\
\hline
VSLNet$^\dagger$~\cite{zhang2020vslnet} 
& 28.15 & 39.07 & 27.59 & 16.65
& 47.33 & 67.26 & 50.46 & 31.53
& 42.26 & 57.75 & 41.10 & 25.58 \\
\textbf{\method{abbr}} (DET) 
& \best{35.49} & \best{50.46} & \best{37.82} & \best{22.54}
& \best{53.09} & {72.50} & 58.33 & \scnd{39.25} 
& \scnd{45.59} & \scnd{64.13} & {44.81} & {26.07} \\
\hline\hline
VSLNet$^{\dagger\star}$
& 30.61 & 41.14 & 30.09 & 18.97
& 53.88 & 71.59 & 57.98 & 41.64
& 49.49 & 65.83 & 49.68 & 32.00 \\
\textbf{\method{abbr}}$^\star$ (ELA) 
& {48.36} & {63.31} & {52.49} & {37.02}
& {62.16} & {79.73} & {69.22} & {51.40} 
& {56.25} & {73.72} & {58.65} & {40.74} \\
\hline
\end{tabular}
\end{center}
\end{table}

\subsection{Ablation Study}
We conducted comprehensive ablation studies
based on the \method{abbr}'s determined predictions
to provide in-depth analyses and better understandings.

\begin{figure}[t]
\begin{minipage}[t]{0.48\linewidth}
\centering
\includegraphics[width=\textwidth]{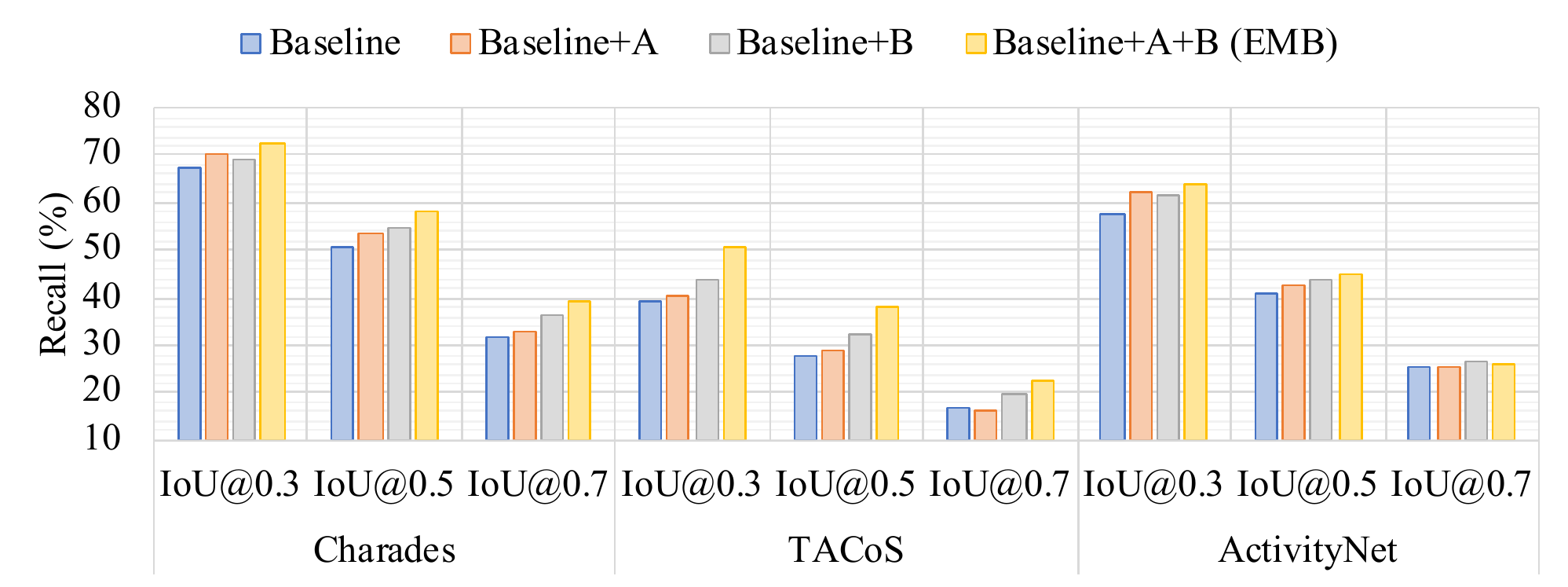}
\caption{Effectiveness of different proposed components.
The elastic moment bounding formulation is denoted as component ``A''
while the multi-grained interaction network is component ``B''.
}
\label{fig:components}
\end{minipage}
\hfill
\begin{minipage}[t]{0.48\linewidth}
\centering
\includegraphics[width=\textwidth]{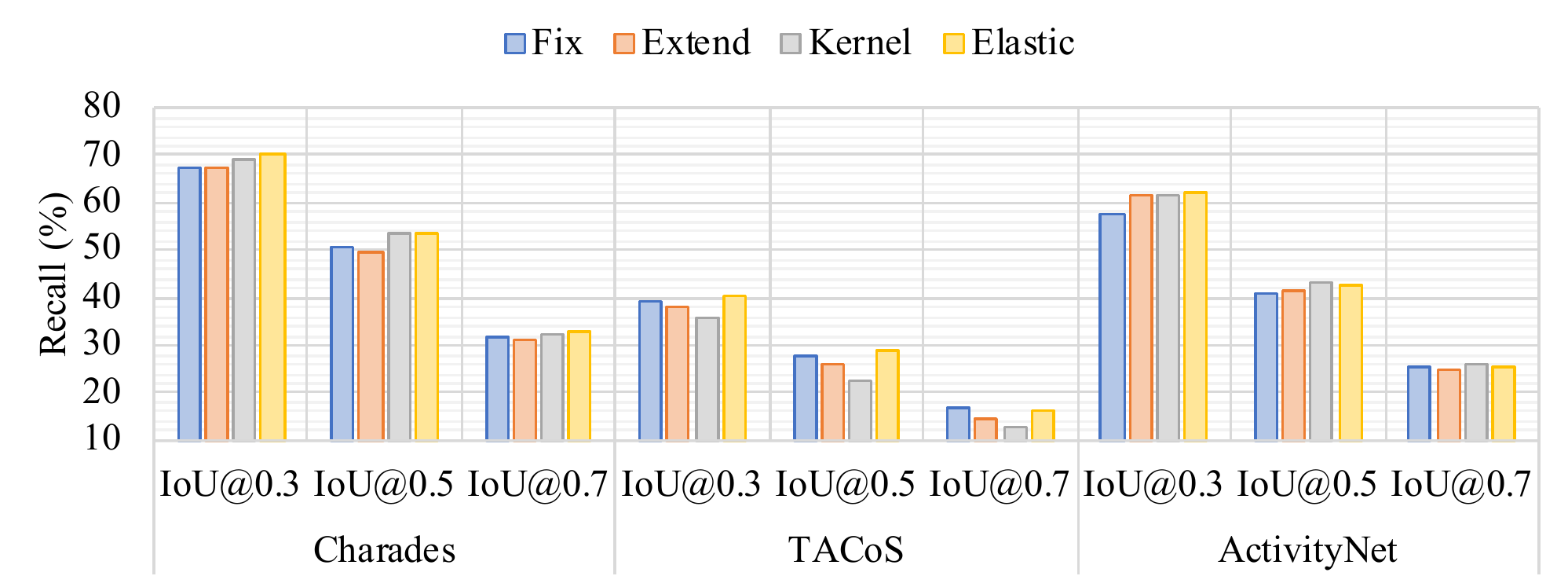}
\caption{Effects of multi-candidate mining strategies in training.
``Fix'': single-frame boundaries.
Multiple endpoints
are generated by extension (Extend), 
a gaussian kernel (Kernel), 
or our elastic bounding.
}
\label{fig:boundary}
\end{minipage}
\end{figure}

\keypoint{Components analysis.}
We investigated the individual contributions of
different components in our \method{abbr} model to 
its improvements over the baseline model~\cite{zhang2020vslnet}.
As shown in Fig.~\ref{fig:components},
both our elastic boundary learning objective (Eq.~\eqref{eq:loss_bound})
and the multi-grained interaction network
brought clear benefits to the baseline.
Such results demonstrate
the effectiveness to learn the temporal endpoints
of video activities with higher flexibility
so to tolerant the uncertainty of manual labels.
Besides,
they also imply the superiority
of our visual encoders which
conduct both within and cross-modal attention learning
and complement the boundary and content information of video segments mutually.

\keypoint{Candidate endpoints mining.}
We evaluated the advantages of 
mining candidate endpoints adaptively
over several heuristic strategies
without the MGIN design:
(1) boundary extension~\cite{zhang2020vslnet},
(2) smoothing by a gaussian kernel~\cite{wang2021smin,xiao2021bpnet} and
(3) single-frame endpoints (baseline).
As shown in Fig.~\ref{fig:boundary},
simply improving the boundary's flexibility 
without considering their reliability (``Extend'')
tends to degrade the model's performances on both datasets.
Boundary smoothing by a gaussian kernel (``Kernel'') 
is sometimes beneficial
but less stable than our adaptive designs
because their candidates
were determined
according to only the duration of MoIs
without considering the video context and query's unambiguity.

\begin{figure}[ht]
\begin{minipage}[t]{0.48\linewidth}
\centering
\includegraphics[width=\textwidth]{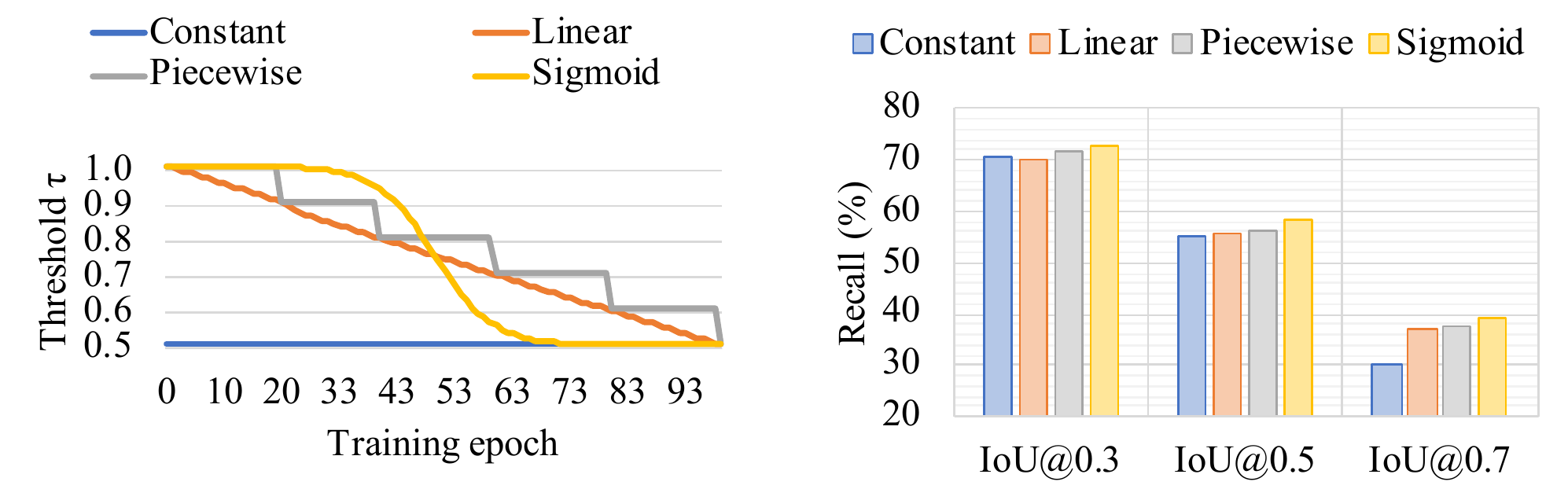}
\caption{Effects of constructing elastic boundary 
subject to an evolving threshold
on Charades-STA.
}
\label{fig:threshold}
\end{minipage}
\hfill
\begin{minipage}[t]{0.48\linewidth}
\centering
\includegraphics[width=\textwidth]{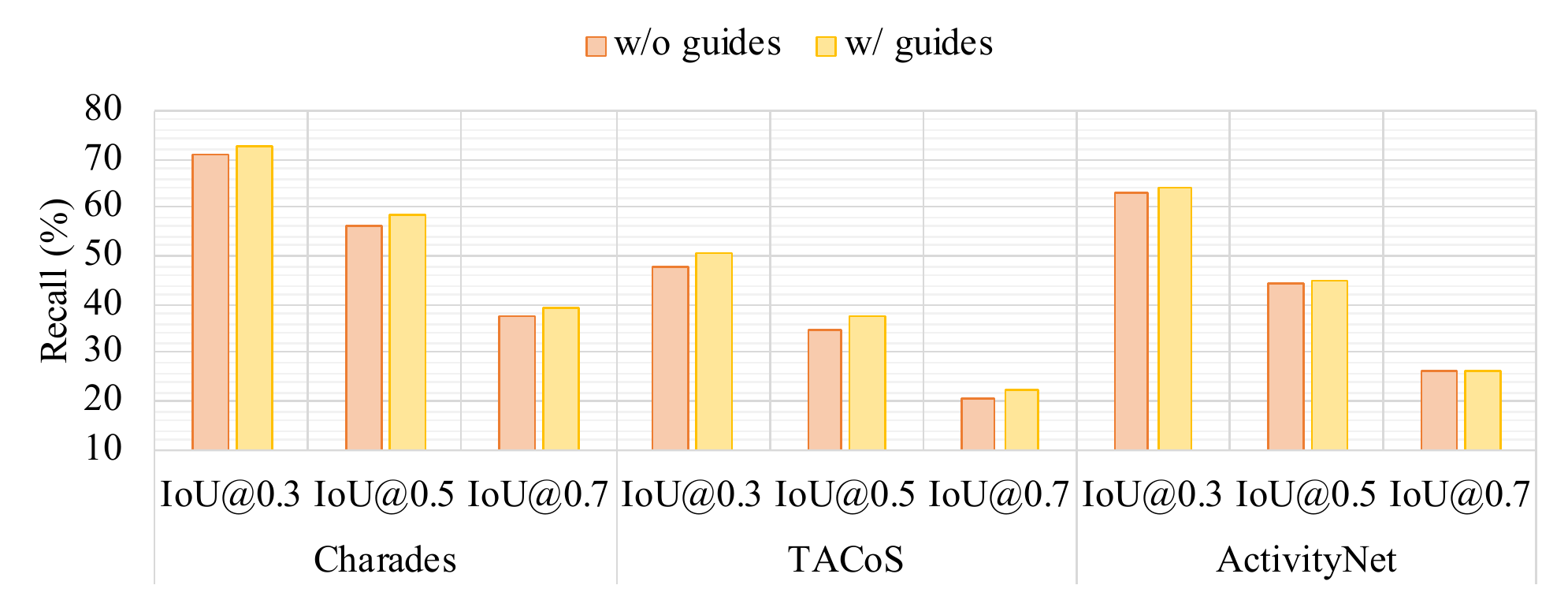}
\caption{Effectiveness of guided attention mechanism
by comparing with the conventional attention modules~\cite{vaswani2017attention}.}
\label{fig:guided}
\end{minipage}
\end{figure}

\keypoint{Evolving threshold.}
We studied the effects of threshold's evolving schemes
to our elastic boundary constructions (Eq.~\eqref{eq:pseudo}).
Fig.~\ref{fig:threshold}
shows the curves of schemes 
and their corresponding performances. 
The model trained with a constant threshold
yielded the worst results
in most cases
while the `Sigmoid' scheme is always the best.
This is because
the `Sigmoid' scheme
maintains a persistently high threshold
at the early training stages
to avoid introducing distractions
when the alignment branch is under-trained,
then drops rapidly
to involve more diverse candidate endpoints
when the alignment branch is reliable.

\keypoint{Guided attention.}
We validated the effectiveness of our guided attention mechanism
by replacing it in our MGIN encoder
by the conventional attention modules proposed in~\cite{vaswani2017attention}.
From the comparison results shown in Fig.~\ref{fig:guided},
the models trained with guided attention outperformed
their counterparts which learned the video representations
without interacting information in multiple granularities.
Such results imply 
the complementary of segment's content and boundary information,
which encourage the video feature representations
to be sensitive to redundancy and activity transitions.

\begin{figure}
\centering
\includegraphics[width=1.0\linewidth]{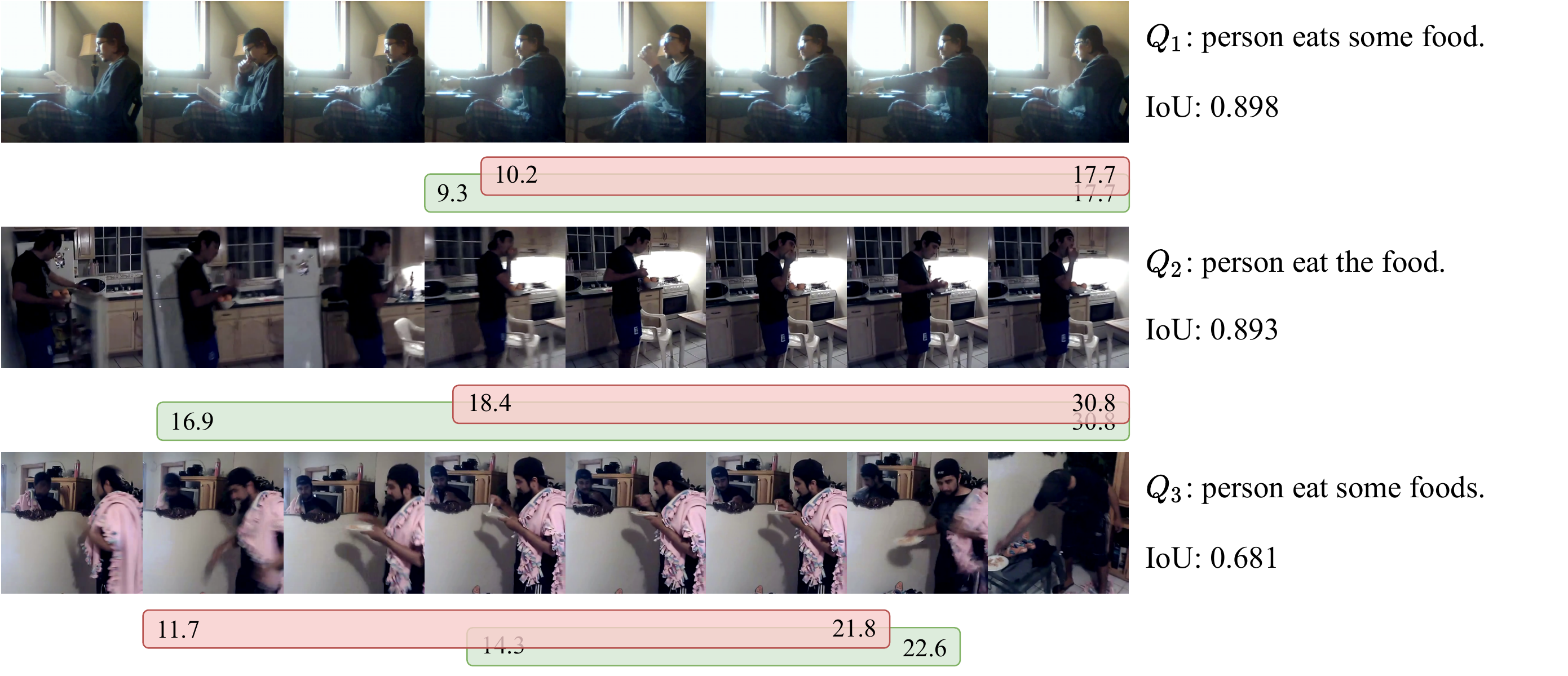}
\caption{
Cases of video activities 
with similar semantics 
but inconsistent manual boundaries.
The manual and predicted boundary 
are shown in red and green, respectively.
}
\label{fig:cases}
\end{figure}

\keypoint{Qualitative case study.}
We provide several video examples in Fig.~\ref{fig:cases}
which are showing video activities
corresponding to semantically similar sentence descriptions.
However,
their manual boundary are inconsistent, 
demonstrating the uncertainty in temporal boundaries.
Specifically,
the manual boundary for $Q_1$
starts from grabbing the food right before putting it into the mouth
while $Q_2$ 
skipping the action of ``grabbing'' and 
starts when the person takes a bite.
The $Q_3$ involves even more redundancy
which covers the actions of taking a plate from a desk and
blending foods by a folk.
In contrast,
the predictions made by our model are more consistent 
on interpreting the action of ``eat'' in different videos,
\ie, always starts from delivering food to the mouth.
This is accomplished by learning with highly flexible boundaries
instead of fitting rigid and ambiguous manual endpoints
which are prone to visual-textual miscorrelations.


\section{Conclusion}
\label{sec:conclude}
In this work,
we introduced a new \method{full} (\method{abbr}) approach
to learn a more robust model for identifying video activity
temporal endpoints with the inherent uncertainty in training labels.
\method{abbr} is based on modelling elastic boundary tailored
to optimise learning more flexibly the endpoints of every target
activity moment with the knowledge that the given training
labels are uncertain with inconsistency.
\method{abbr} learns a more accurate and robust visual-textual
correlation generalisable to activity moment localisation in more
naturally prolonged unseen videos where activity of interests are
fractionally small and harder to detect.
Comparative evaluations and ablation studies 
on three activity localisation benchmark datasets
demonstrate the competitiveness and unique advantages of
\method{abbr} over the state-of-the-art models especially
when the untrimmed videos are long and activity moments are short.


\section*{Acknowledgements}
This work was supported by the China Scholarship Council, Vision Semantics Limited, 
the Alan Turing Institute Turing Fellowship, Adobe Research and Zhejiang Lab (NO.
2022NB0AB05).

\clearpage
%
%
\bibliographystyle{splncs04}
\bibliography{press,egbib}
\end{document}